\pdfoutput=1
\documentclass[11pt]{article}

\usepackage[]{ACL2023}

\usepackage{times}
\usepackage{latexsym}
\usepackage{svg}

\usepackage{array}
\usepackage{amssymb}
\usepackage{amsmath}
\usepackage{algorithm}
\usepackage{algpseudocode}
\usepackage{multirow}
\usepackage{algorithmicx}

\usepackage[T1]{fontenc}
\usepackage[utf8]{inputenc}

\usepackage{microtype}

\usepackage{inconsolata}

\usepackage{subfig}
\usepackage{pgfplots}
\usepgfplotslibrary{fillbetween}
\usetikzlibrary{patterns}
\usetikzlibrary{pgfplots.groupplots}
\pgfplotsset{compat=1.17}
\usepackage{lscape}
\usepackage{multirow}
\usepackage{dblfloatfix}
\usepackage{footnote}
\usepackage{graphicx}
\usepackage{tikz}
\usepackage{booktabs, tabularx}
\usepackage{amssymb}
\usepackage{xcolor}

\title{LM-CPPF: Paraphrasing-Guided Data Augmentation \\ 
for Contrastive Prompt-Based Few-Shot Fine-Tuning}

\author{
  Amirhossein Abaskohi\textsuperscript{1}, Sascha Rothe\textsuperscript{2}, Yadollah Yaghoobzadeh\textsuperscript{1,3} \\
  \textsuperscript{1}School of Electrical and Computer Engineering \\ College of Engineering, University of Tehran, Tehran, Iran \\
  \textsuperscript{2}Google DeepMind, Zürich, Switzerland \\
  \textsuperscript{3} Tehran Institute for Advanced Studies, Khatam University, Iran\\
\normalsize{\texttt{amir.abaskohi@ut.ac.ir, rothe@google.com, y.yaghoobzadeh@ut.ac.ir}}
}

\begin{document}
\maketitle

\begin{abstract}

In recent years, there has been significant progress in developing pre-trained language models for NLP. However, these models often struggle when fine-tuned on small datasets. To address this issue, researchers have proposed various adaptation approaches. Prompt-based tuning is arguably the most common way, especially for larger models. Previous research shows that adding contrastive learning to prompt-based fine-tuning is effective as it helps the model generate embeddings that are more distinguishable between classes, and it can also be more sample-efficient as the model learns from positive and negative examples simultaneously. One of the most important components of contrastive learning is data augmentation, but unlike computer vision, effective data augmentation for NLP is still challenging. This paper proposes LM-CPPF, Contrastive Paraphrasing-guided Prompt-based Fine-tuning of Language Models, which leverages prompt-based few-shot paraphrasing using generative language models, especially large language models such as GPT-3 and OPT-175B, for data augmentation. Our experiments on multiple text classification benchmarks show that this augmentation method outperforms other methods, such as easy data augmentation, back translation, and multiple templates.\footnote{Our implementation is publicly available at:  \url{https://github.com/AmirAbaskohi/LM-CPPF}}

\end{abstract}

\section{Introduction}
\label{sec:introduction}

Pre-trained language models (PLMs) are trained on large-scaled corpora in a self-supervised fashion. They have fundamentally changed the NLP community in the past few years by achieving impressive results in various Tasks \cite{devlin2018bert, radford2018improving, yang2019xlnet, chiang-etal-2022-recent}. However, when PLMs are fine-tuned on small datasets, their performance declines. Researchers have proposed various techniques to adapt PLMs to these scenarios \cite{snell2017prototypical, sung2018learning}. In addition to performance, fine-tuning PLMs to learn a new task is parameter inefficient, because an entirely new model is required for every task \cite{houlsby2019parameter}. 

By the introduction of GPT-3 \cite{https://doi.org/10.48550/arxiv.2005.14165} with 175B parameters, it has been shown that Large Language Models (LLMs) are efficient few-shot learners as they can use their knowledge more effectively. One of the key features of these LLMs is their ability to perform multiple tasks using prompts. A language prompt is a piece of text that is added to the input query to help the model make more accurate predictions. In addition, LLMs can be fine-tuned for specific tasks using few examples. This has made them powerful tools for NLP tasks, especially in few-shot scenarios. However, that might not be practical for many situations because of the model size. Therefore, there is a need to adapt smaller PLMs to work in a similar way to LLMs.

Prompt-based fine-tuning is a method for adapting PLMs to specific tasks or domains by providing a prompt \cite{schick2020exploiting, schick2020s}. This approach has been shown to be effective in various NLP tasks, including text classification \cite{https://doi.org/10.48550/arxiv.2105.11259, https://doi.org/10.48550/arxiv.2205.05313} and question answering \cite{yao2022prompt}. However, it can be challenging to achieve strong performance when only a few examples are available for each task. \newcite{gao2020making} introduced a prompt-based fine-tuning method called LM-BFF for RoBERTa \cite{liu2019roberta} to tackle this issue. Their approach includes automated prompt generation and a more effective way of using task examples in fine-tuning.

Building on the success of LM-BFF and considering contrastive learning's promising results both in computer vision \cite{chen2020simple} and NLP \cite{chen2020simple, miao2021simple}, \newcite{jian-etal-2022-contrastive} present a contrastive learning framework to improve LM-BFF. They propose a Supervised Contrastive Learning (SCL) approach \cite{khosla2020supervised} that classifies inputs using different augmented views of the data. These views are created using different templates for their demonstrations when building prompts.

In this paper, we show that while SCL at the feature space can be beneficial, the use of different templates can limit the full potential of this approach. We propose \textbf{LM-CPPF} (Contrastive Paraphrasing-guided Prompt-based Fine-tuning of Language Models), in which we integrate the knowledge of LLMs like GPT-3 and OPT-175B \cite{zhang2022opt} to build different views using paraphrasing. These models can generate paraphrases of a sentence with different syntax, not just by changing the lexicalization. Previous studies have considered generating paraphrases a challenging and costly NLP task \cite{siddique2020unsupervised, garg2021unsupervised, zhou2021paraphrase}. However, PLMs can generate paraphrases easily and effectively using in-context learning with few examples. Although prior research has studied paraphrase generation with PLMs \cite{roy2019unsupervised, hegde2020unsupervised}, to the best of our knowledge, this is the first time that large LLMs are utilized to generate paraphrases with prompts as an augmentation method. Our experiments on six different text classification tasks demonstrate that LM-CPPF outperforms the previous SOTA methods of data augmentation in prompt-based fine-tuning, including Easy Data Augmentation (EDA) \cite{wei2019eda}, Back Translation (BT) \cite{sugiyama2019data}, and multiple templates \cite{jian-etal-2022-contrastive}.

\section{Related Works}
\label{sec:realted-works}

LLMs like GPT-3 \cite{brown2020language} can perform NLP tasks with few examples and natural prompts. But smaller models are not efficient with this approach and there are data sparsity and prompt sensitivity issues. To address these challenges, \newcite{gao2021making} propose LM-BFF, a framework that leverages a large PLM to automatically generate task-specific prompts for smaller models. It improves their few-shot performance on different NLP tasks. Some work have enhanced LM-BFF with different prompt tuning methods. For example, \newcite{zhou-etal-2022-dual} present a dual context-guided continuous prompt tuning method that uses the language context and connects discrete and continuous prompt tuning. \newcite{jian-etal-2022-contrastive} integrate contrastive learning and data augmentation with LM-BFF. In their contrastive part, in addition to comparing different instances from the same or different classes, they introduced a novel prompt-specific augmentation method. In their approach, they change the template of the prompt. In this paper, we use few-shot paraphrasing with LLMs for contrastive prompt-tuning, which fine-tunes models with natural prompts.

Paraphrasing is the task of expressing the same meaning with different words or structures. It can be used to create training data with increased diversity and naturalness for NLP tasks, such as text classification \cite{xie2020unsupervised}, natural language inference \cite{kumar-etal-2019-submodular}, and text summarization \cite{loem-etal-2022-extraphrase}, surpassing the limitations of traditional approaches. Paraphrasing helps with data scarcity and model generalization. There are different ways to generate paraphrases for data augmentation. One is back-translation \cite{sennrich-etal-2016-improving}, which uses a translation system to convert a sentence to another language and back. Another is to use paraphrasing models trained on parallel paraphrase datasets \cite{wieting-gimpel-2018-paranmt, zhu-etal-2022-duqm}. PLMs can also generate paraphrases by using large-scale corpora, but they may produce paraphrases that are not semantically consistent or relevant. LLMs can reduce this problem as they encode and generate language better. In this paper, we generate paraphrases by carefully prompting LLMs and then use them for data augmentation.

\section{Method}
\label{sec:method}

\begin{figure*}
  \centering
  \includegraphics[width=16cm, keepaspectratio]{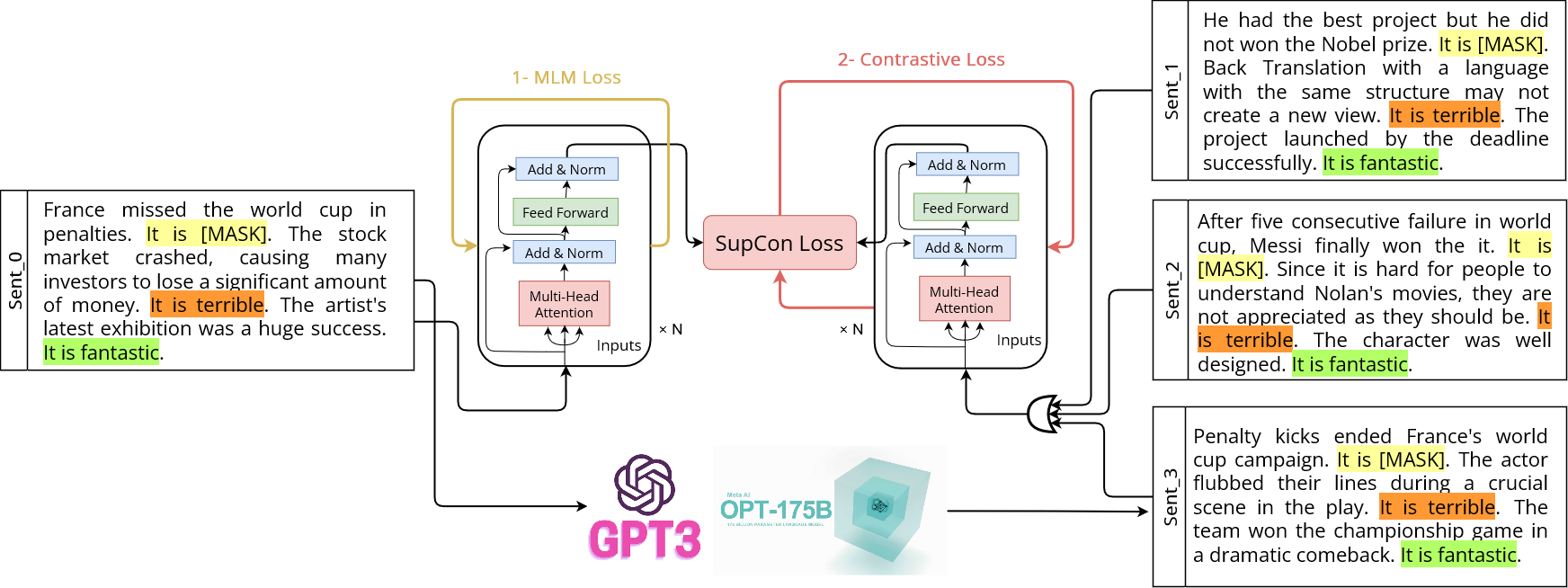}
  \caption{\label{fig:fine-tuning} Our method, LM-CPPF, includes two objectives: (I) MLM and (II) Supervised Contrastive Learning. The target sentence is the first sentence in each prompt with a [MASK] token. The target sentence of Sent\_0 is used to train our model and calculate the MLM loss. We build Sent\_3, whose target sentence is a paraphrase of Sent\_0's target sentence. Sent\_1 and Sent\_2, sampled from the dataset, have target sentences in the same and different classes as Sent\_0, respectively.}
%   By comparing the embeddings of Sent\_1, Sent\_2, or Sent\_3's target sentence with Sent\_0's target sentence, we calculate the SupCon loss and run the backward process for the second time. }
\end{figure*}
\paragraph{Background}
Contrastive learning's success relies on data augmentation, which creates new views of the input data. Contrastive learning has been utilized for various tasks in deep learning \cite{le2020contrastive,conde2021clip,abaskohi2022automatic}; however, most NLP data augmentation methods may influence semantics which results in limited improvement. For instance, EDA's synonym substitution may create entirely new samples since words do not have equal senses \cite{keselj2009speech}. In addition to these augmentation methods, the approach used in \newcite{jian-etal-2022-contrastive} cannot be counted as data augmentation as the sample is still the same and only the template for the verbalizer changes. Although it is a creative approach designed specifically for the prompt-based method of LM-BFF, it is limited in performance even compared to EDA in several benchmarks. Furthermore, it requires an expert to create multiple templates for each task, which makes it challenging for newly emerged tasks. Here we propose leveraging LLMs to generate paraphrases and introduce LM-CPPF, a novel approach aimed at addressing the challenges associated with contrastive prompt-based fine-tuning of PLMs.

\paragraph{Few-shot paraphrasing}
Paraphrasing is one of the best methods for data augmentation in NLP. One of the most popular approaches for paraphrasing is back-translation (BT) \cite{sugiyama2019data} due to its simplicity and efficiency. Nonetheless, BT's performance depends a lot on the intermediary language. In this paper, we, instead, use a combination of prompt-learning and LLMs for paraphrasing. In few-shot paraphrasing, an LLM rewrites a sentence given an instruction and a few examples. We believe that LLMs generate high-quality paraphrases due to their encoded semantic and sentence structure knowledge. 
% While this method does rely on extra knowledge, BT also relies on the knowledge of another language model. To generate paraphrases, 
We utilize GPT-3 \cite{https://doi.org/10.48550/arxiv.2005.14165} or OPT-175B \cite{zhang2022opt} via their official APIs \footnote{OPT-175B: \url{opt.alpa.ai} and GPT-3: \url{openai.com/api}} for generating paraphrases. 
% Moreover, we compare  GPT-2 \cite{radford2019language} to as a smaller LM

To avoid violating the prompt-based fine-tuning settings, we do not include any additional task data in generating our paraphrases. Following the few-shot setting in LM-BFF, we assume to have access to a PLM $M$, datasets $\mathcal{D}_{train}$, and $\mathcal{D}_{test}$ with label space $\mathcal{Y}$ where there are only $\mathcal{K} = 16$ examples per class in $\mathcal{D}_{train}$. We use this setting for both prompt-based few-shot paraphrasing and fine-tuning. To generate paraphrases, excluding the one sample that we want to paraphrase, we use QuillBot\footnote{\url{quillbot.com}} to create paraphrases for our prompts for the remaining 15 samples in the same class of $\mathcal{D}_{train}$. We leverage two types of prompts for paraphrasing:
    (I) \textbf{Only Demonstration:} Here, the samples and their paraphrased versions are given using the templates in Table \ref{table:onlyexamples} to demonstrate the task of paraphrasing. 
    (II) \textbf{Demonstrations with Instruction:} In addition to the previous method, this one includes instructions at the beginning of the prompt, defining paraphrasing before demonstrations. These instructions can be seen in Table \ref{table:examplewithinstruction}.

\begin{table*}[]
    % \small
    \centering
    \tabcolsep=0.14cm
    \begin{tabular}{c|ccc|cc|cc}
    \hline
    \multirow{2}{*}{Task} & \multirow{2}{*}{LM-BFF} & LM-BFF+    & LM-BFF+                 & LM-CPPF       & LM-CPPF & LM-CPPF & LM-CPPF          \\
                          &                         & SupConLoss & Multi-templates & GPT-3         & OPT     & GPT-2   & FT GPT-2 \\ \hline
    SST-2                 & 89.5                    & 90.3       & 91.0                    & \textbf{92.3} & 91.8    & 91.1    & 91.4             \\
    SST-5                 & 48.5                    & 49.6       & 50.3                    & \textbf{52.8} & 52.2    & 51.4    & 51.6             \\
    MNLI                  & 62.3                    & 63.2       & 64.8                    & \textbf{68.4} & 66.2    & 65.6    & 65.8             \\
    CoLA                  & 6.9                     & 9.6        & 11.6                    & \textbf{14.1} & 13.3    & 10.7    & 11.8             \\
    QNLI                  & 61.2                    & 65.4       & 67.2                    & \textbf{69.2} & 68.5    & 67.5    & 67.8             \\
    CR                    & 89.7                    & 89.9       & 90.2                    & \textbf{91.4} & 91.1    & 90.2    & 90.7             \\ \hline
    \end{tabular}
    \caption{\label{table:paraphraseFinetuneRes} Performance of LM-CPPF and our baselines in six datasets. LM-BFF+Multi-templates refers to \newcite{jian-etal-2022-contrastive}. LM-BFF+SupConLoss uses the same architecture of LM-BFF+Multi-templates, but without any data augmentation, just integrating supervised contrastive and MLM loss functions. Two cases are available for GPT-2: the pre-trained model and the GPT-2 fine-tuned (FT) on ParaNMT-50M \cite{wieting-gimpel-2018-paranmt} dataset.
    LM-BFF, LM-BFF+Multi-template, and  LM-CPPF (on average for all models used for paraphrasing) have 0.77 and 1.02, and 1.65 standard deviations on average for each task, respectively.}
\end{table*}

\paragraph{Contrastive prompt-based fine-tuning}

LM-CPPF consists of two steps. The first step involves calculating the Masked Language Modeling (MLM) loss by using the target sentence in the given template, the specific demonstrations in the prompt, and the verbalizer matched with the target sentence's label. We calculate the supervised contrastive loss in the second step by comparing the target prompt with another sample with the same template but different random demonstrations. This comparison sample can be in the same or a different class as the target prompt. When the comparison sample belongs to a different class, it is randomly sampled from the dataset. However, in cases where the comparison sample belongs to the same class, an alternative approach is employed. This involves either selecting another sample from the same class within the dataset or applying data augmentation techniques, paraphrasing in our case, to augment the target sample in order to create a new view of it. In both of these cases, the demonstrations are not the same. Figure \ref{fig:fine-tuning} illustrates the fine-tuning process, and Algorithm \ref{alg:mlm-supcon} shows our methodology when paraphrasing creates a new view of the target sample. See Appendix \ref{appendix:contrastive prompt-based fine-tuning details} for more information.

\section{Experiments}
\label{sec:experiments}

\paragraph{Evaluation datasets and protocol} Our method is evaluated on six different classification tasks from LM-BFF \cite{liu2021bootstrapping}. The reported numbers represent the average accuracy from five runs using Roberta-base \cite{liu2019roberta}. In Section \ref{subsec:ParaphrasingInPromptFinetuning} where LLMs are compared for paraphrasing, we also employed pre-trained and fine-tuned GPT-2 as an additional model for paraphrasing, allowing us to leverage smaller models in our experiments. For the fine-tuning of GPT-2 specifically for paraphrasing, we utilized the ParaNMT-50M \cite{wieting-gimpel-2018-paranmt} dataset. More details regarding the training process can be found in Appendix \ref{appendix:BatchSizeAndLearningDetails}.

\subsection{Paraphrasing in Prompt Fine-tuning}
\label{subsec:ParaphrasingInPromptFinetuning}

This section presents the results of our fine-tuning approach using paraphrasing on various NLP tasks. As shown in Table \ref{table:paraphraseFinetuneRes}, LM-CPPF improves the model's accuracy on all tasks compared to the baseline method of LM-BFF+Multi-templates \cite{jian-etal-2022-contrastive}. Comparing the standard deviation of our model in five runs and the standard deviations of LM-BFF and LM-BFF + Multi-templates, we see that LM-CPPF has a higher standard deviation as it uses an intermediary model for generating paraphrases. In contrast, LM-BFF + Multi-templates integrates templates that have nearly equal performance \cite{jian-etal-2022-contrastive}.

We also compare the effect of using GPT-3, OPT-175B, and GPT-2 as our language model for few-shot paraphrasing. We did two experiments with GPT-2 large: (I) Using a pre-trained version of GPT-2 where the weights are not tuned at all (II) Fine-tuned GPT-2 where the model has been fine-tuned on the ParaNMT-50M dataset. The results in Table \ref{table:paraphraseFinetuneRes} indicate that GPT-3 outperforms OPT-175B in all tasks and GPT-2 has a lower performance, which was predictable since it has significantly fewer parameters. Also, fine-tuned GPT-2 shows a better performance which suggests that GPT-2's knowledge after pre-training is not enough for doing a task like paraphrasing. About the LLMs, although both models have 175B parameters, OPT-175B has a $1/7$ carbon footprint of GPT-3, and it is also freely available \cite{zhang2022opt}. Consequently, we base our further analysis on OPT-175B.

\begin{table*}[]
    \centering
    % \fontsize{9.5}{11}\selectfont
    \begin{tabular}{c|c|ccccc|ccccc}
    \hline
        \multirow{2}{*}{Task} & Few-shot     & \multicolumn{5}{c|}{Back Traslation} & \multirow{2}{*}{SR} & \multirow{2}{*}{RI} & \multirow{2}{*}{RS} & \multirow{2}{*}{RD} & \multirow{2}{*}{EDA} \\
                              & Paraphrasing & AR    & FR    & DE    & ZH   & HI & &      &      &      &      \\ \hline
        SST-2                 & \textbf{91.8}         & \underline{90.8}  & 90.6  & 90.4  & 90.7 & 90.3 & 90.5 & 89.5 & 90.8 & \underline{91.3} & 90.4 \\
        SST-5                 & \textbf{52.2}         & 49.2  & 49.3  & 49.1  & \underline{49.6} & 48.3 & 47.9 & \underline{49.3} & \underline{49.3} & 48.2 & 48.2 \\
        MNLI                  & \textbf{66.2}         & 64.3  & 63.1  & 63.8  & \underline{65.4} & 62.2 & 62.9 & \underline{63.2} & 61.7 & 60.2 & 60.3 \\
        CoLA                  & \textbf{13.3}         & 6.7   & 6.8   & 6.4   & \underline{7.1}  & 5.9  & \underline{6.3}  & 5.8  & 5.8  & 5.1  & 5.1  \\
        QNLI                  & \textbf{68.5}         & 66.5  & 66.2  & 65.8  & \underline{66.6} & 64.3 & 66.1 & 65.9 & \underline{66.3} & 65.6 & 63.3 \\
        CR                    & \textbf{91.1}         & 88.5  & 88.6  & 88.4  & \underline{88.7} & 87.9 & \underline{89.8} & 89.1 & 89.3 & 89.6 & 89.7 \\ \hline
    \end{tabular}
    \caption{\label{table:ParaphrasingVTranslation} Comparing the accuracy of our few-shot paraphrasing approach with the Back Translation (BT) and Easy Data Augmentation (EDA) methods. EDA includes Synonym Replacement (SR), Random Insertion (RI), Random Swap (RS), and Random Deletion (RD). EDA in the results is combination of all of the four mentioned methods. BT and EDA standard deviations are 1.31 and 1.4 on average, respectively, while our approach has a standard deviation of 1.65.}
\end{table*}

\subsection{Few-shot Paraphrasing vs. Other Data Augmentation Methods}
\label{subsec:ParaphrasingVTranslation}

\begin{table}[]
    \small
    \centering
    \begin{tabular}{c|cccccc}
        \toprule
        \multirow{2}{*}{\textbf{Task}} & \multicolumn{6}{c}{\textbf{Template Number}}                       \\
                              & 1    & 2    & 3             & 4    & 5    & 6             \\ \hline
        SST-2                 & 91.8 & 91.2 & 91.4          & 89.1 & 92.1 & \textbf{92.4} \\
        SST-5                 & 52.2 & 53.1 & 52.7          & 53.4 & 53.6 & \textbf{54.1} \\
        MNLI                  & 66.2 & 65.9 & \textbf{66.9} & 66.1 & 66.2 & 66.4          \\
        CoLA                  & 13.3 & 12.7 & 13.2          & 13.8 & 13.4 & \textbf{13.6} \\
        QNLI                  & 68.5 & 68.4 & 68.6          & 68.5 & 68.8 & \textbf{69.3} \\
        CR                    & 91.1 & 91.2 & 91.3          & 91.5 & 91.7 & \textbf{92.2} \\ \bottomrule
    \end{tabular}
    \caption{Performance of different paraphrasing prompt demonstration templates.
    \label{table:differentPromptTemplates}}
\end{table}

In this section, we present an experimental comparison of the performance of the few-shot paraphrasing approach and other data augmentation methods, including BT and EDA. The results are shown in Table \ref{table:ParaphrasingVTranslation}. The BT approach is evaluated using different intermediary languages (Arabic, French, Deutsch, Chinese, and Hindi). The results indicate that BT's performance is slightly different across languages, with Chinese showing the highest performance. In general, paraphrasing approaches, including BT, are better in comparison to EDA. In SST-2 and CR, where the samples are usually simple sentences, BT shows weaker performance than EDA. We believe the reason is that BT can be more effective for longer sequences because longer sequences usually contain more context and nuanced meaning. Moreover, EDA employs additional knowledge from another PLM in certain actions, such as synonym substitution, similar to BT and few-shot paraphrasing. 

The few-shot paraphrasing approach introduced in this work outperforms both BT and EDA. This confirms that using PLM's knowledge properly in paraphrasing is an effective and efficient data augmentation method. In few-shot paraphrasing, we instruct the model to generate paraphrases that differ in lexicalization and sentence structure. 

% Moreover, as BT adds the loss of two forwarding phases (from source to intermediary and from intermediary to the source language), and just substitutes few words with little changes in the structure of the sentence, it has poorer performance than few-shot paraphrasing. 

\subsection{Prompt Template Evaluation}
\label{subsec:templateEvaluation}
As the heart of our method is the few-shot paraphrase generation done 
by LLMs, we investigate the impact of different paraphrasing prompt demonstrations and instruction templates on the performance of our model. Table \ref{table:differentPromptTemplates} shows that the last template presented in Table \ref{table:onlyexamples} is better in almost all tasks. 
This template, ``<Original Text>, in other words <Paraphrased>'', uses a complete and concrete sentence, unlike other templates, which use specific tokens, such as ``[Original]'', to distinguish between the original and the paraphrased version. 
Also, we compare different instruction templates presented in Table \ref{table:examplewithinstruction}. As we aimed to report our best result in each task here, we used the best demonstration template for any particular task, which was determined in Table \ref{table:differentPromptTemplates}. Table \ref{table:promptInstructionTemplate} shows that the fourth template achieves the best performance, as it precisely describes the task with its instruction ``Generate a paraphrase of the following text using different words and sentence structures while still conveying the same meaning''.

\begin{table}[]
    \small
    \tabcolsep=0.14cm
    \centering
    \begin{tabular}{c|c|ccccc}
    \toprule
        \multirow{2}{*}{\textbf{Task}} & \multirow{2}{*}{w/o Instruct} & \multicolumn{5}{c}{\textbf{Template Number}}       \\
                              &                                      & 1    & 2    & 3    & 4             & 5    \\ \hline
        SST-2                 & 92.4                                 & 93.1 & 93   & 92.8 & \textbf{93.2} & 92.7 \\
        SST-5                 & 54.1                                 & 54.7 & 54.5 & 54.2 & \textbf{54.9} & 54.3 \\
        MNLI                  & 66.9                                 & 67.8 & 67.5 & 67.1 & \textbf{68.2} & 67.2 \\
        CoLA                  & 13.6                                 & 13.1 & 13.2 & 12.6 & \textbf{13.3} & 12.8 \\
        QNLI                  & 69.3                                 & 69.8 & 70.1 & 69.5 & \textbf{70.2} & 69.6 \\
        CR                    & 92.2                                 & 93.1 & 92.8 & 92.6 & \textbf{93.3} & 92.4 \\ \bottomrule
    \end{tabular}
    \caption{Performance of different paraphrasing prompt instruction templates on various NLP tasks. 
    \label{table:promptInstructionTemplate}}
\end{table}

% \subsection{GPT-3 vs. OPT-175B}
% \label{subsec:gpt3VOPT175B}

% \begin{table}[]
%     \centering
%     \caption{The results, shown in the table, indicate that GPT-3 outperforms OPT-175B on all tasks, with the greatest increase in accuracy seen on the CoLA task.\label{table:gpt3VOPT}}
%     \begin{tabular}{ccc}
%         \hline
%         Task  & GPT-3 & OPT-175B \\ \hline
%         SST-2 & 92.3  & 91.8     \\
%         SST-5 & 52.8  & 52.2     \\
%         MNLI  & 68.4  & 66.2     \\
%         CoLA  & 14.1  & 13.3     \\
%         QNLI  & 69.2  & 68.5     \\
%         CR    & 91.4  & 91.1     \\ \hline
%     \end{tabular}
% \end{table}

\section{Conclusion}
\label{sec:conclusion}

Our experiments demonstrated the effectiveness of using few-shot paraphrasing as a data augmentation method for contrastive prompt-based fine-tuning of PLMs. It outperformed other data augmentation methods in text classification tasks, such as EDA, multiple templates, and back translation. We also found that our approach is effective with GPT-3 or OPT-175b models in generating paraphrases. Overall, LM-CPPF improves the performance of LM-BFF by large margins using contrastive learning applied on paraphrases generated by LLMs.

\section*{Limitations}
Our approach relies on the performance of the few-shot paraphrasing. This results in two limitations for our approach. One limitation is the difficulty in accessing GPT-3 and OPT-175b models. These models currently need to be more widely available. OPT-175B has a free version but it is very slow. Another limitation is the need for annotated demonstrations for few-shot paraphrasing. While there are available models and tools, like QuillBot, that can be used for this purpose,  their quality is not comparable to GPT-3 and OPT-175b. This can limit the power of these tools in our approach. Using human knowledge to paraphrase the demonstration can help these large models generate high-quality paraphrases but it is expensive. 

\section*{Ethics Statement}
The research conducted in this paper has been carried out in accordance with the ethical principles of ACL. We have ensured that our experiments do not harm any individuals or groups and have obtained informed consent from all participants. As mentioned in the paper, we also tried to base our main experimentation on the more environmentally-friendly option,  OPT-175B. 

\bibliography{anthology,custom}
\bibliographystyle{acl_natbib}

\appendix

\renewcommand{\thetable}{\Alph{section}.\arabic{table}}
\setcounter{table}{0}

\renewcommand{\thealgorithm}{\Alph{section}.\arabic{algorithm}}

\section{Evaluation Setting}
\label{appendix:BatchSizeAndLearningDetails}

We used a learning rate of $1e^{-5}$ for MLM loss like LM-BFF. Although contrastive learning algorithms often perform better with larger batch training, due to resource limitations, we had to use half the batch size suggested in \newcite{jian-etal-2022-contrastive} for various tasks in the SCL phase. As recommended in \newcite{krizhevsky2014one}, we used $sqrt(0.5) \approx 0.7$ of the learning rates mentioned in \newcite{jian-etal-2022-contrastive} for this phase. Therefore, we report baselines with our smaller batch size. Our method uses a single template for each task's prediction. The primary prompts are listed in Appendix \ref{appendix:TaskPrompt}. For the prompts used in the paraphrasing phase, with the exception of experiments in Section \ref{subsec:templateEvaluation}, we used randomly selected templates from the suggested prompts listed in Table \ref{table:onlyexamples}. In all of the experiments, we used OPT-175B, except one of the results mentioned in Section \ref{subsec:ParaphrasingInPromptFinetuning}, where we compared OPT-175B and GPT-3 in paraphrasing.

We show the batch size and learning rate for SupCon in Table \ref{table:batchsizeandlearningrate}. It is important to note that the results of LM-BFF presented in the main paper were obtained using the same large batch size as our method to ensure fair comparisons.

We fine-tuned with a batch size that fits into GPU memory and is divisible by the total number of examples in the task. Experiments were conducted on one NVIDIA RTX-3090 with 24 GB memory using the RoBERTa-base model. Furthermore, as per LM-BFF, we fine-tuned for a maximum of 1000 steps.

\begin{table}[!ht]
    \centering
    \begin{tabular}{c|cc}
    \hline
    Task  & Batch Size & Learning Rate \\ \hline
    SST-2 & 8          & $7e^{-7}$     \\
    SST-5 & 20         & $7e^{-6}$     \\
    MNLI  & 12         & $7e^{-6}$     \\
    CoLA  & 8          & $7e^{-6}$     \\
    QNLI  & 8          & $7e^{-6}$     \\
    CR    & 16         & $7e^{-6}$     \\ \hline
    \end{tabular}
    \caption{Batch size and learning rate for SupCon loss used for each task.}
    \label{table:batchsizeandlearningrate}
\end{table}

For the GPT-2 experiments in Table \ref{table:paraphraseFinetuneRes}, we followed the same intructions for generating paraphrases as we used for GPT-3 and OPT-175. In fine-tuning GPT-2, we fine-tuned our model on ParaNMT-50M \cite{wieting-gimpel-2018-paranmt} with the batch size of 32 and learning rate of $1e^{-3}$ for 5 epochs.

\section{Task Prompts}
\label{appendix:TaskPrompt}

The primary prompts utilized for each task in our experiments are displayed in Table \ref{table:primarytemplates}. They were handpicked by LM-BFF \cite{gao2021making}.

\begin{table*}[!ht]
    \centering
    \small
    \begin{tabular}{c|cc}
    \hline
    Task  & Template                                                             & Verbalizers                                                                           \\ \hline
    SST-2 & \textless{}S1\textgreater It was {[}MASK{]} .                        & positive: great, negative: terrible                                                   \\
    SST-5 & \textless{}S1\textgreater It was {[}MASK{]} .                        & v.positive: great, positive: good, neutral: okay, negative: bad, v.negative: terrible \\
    MNLI  & \textless{}S1\textgreater ? {[}MASK{]} , \textless{}S2\textgreater{} & entailment: Yes, netural: Maybe, contradiction: No                                    \\
    CoLA  & \textless{}S1\textgreater This is {[}MASK{]} .                       & grammatical: correct, not\_grammatical: incorrect                                     \\
    QNLI  & \textless{}S1\textgreater ? {[}MASK{]} , \textless{}S2\textgreater{} & entailment: Yes, not\_entailment: No                                                  \\
    CR    & \textless{}S1\textgreater It was {[}MASK{]} .                        & positive: great, negative: terrible                                                   \\ \hline
    \end{tabular}
    \caption{Primary templates and verbalizers (label words) used in our experiments.}
    \label{table:primarytemplates}
\end{table*}

\section{Paraphrasing Prompts}
\label{appendix:Paraphrasing Prompts}

To find the best prompt for paraphrasing, we checked different corpus available online and found out how the paraphrasing examples are introduced. We generated our prompts by using this information and our manual modification in these templates.

In this demonstration prompt, we did not provide any explanations or descriptions for the specific transformation applied to the input to produce the output. Instead, we labeled the original sample and its paraphrase. For instance, we used the token \textbf{[Original]} to indicate the original sentence in the dataset and the token \textbf{[Paraphrase]} to indicate the paraphrased sample. Table \ref{table:onlyexamples} shows the templates we used for this approach.

\begin{table}[!ht]
    \centering
    \begin{tabular}{>{\centering\arraybackslash}p{3in}}
    \textbf{Demonstration Template}                               \\ \hline
    Original:\textless{}Original Text\textgreater{} \\ Paraphrase:\textless{}Paraphrased Text\textgreater{} \\ \hline
    \lbrack Original\rbrack:\textless{}Original Text\textgreater{} \\ \lbrack Paraphrase\rbrack:\textless{}Paraphrased Text\textgreater{} \\ \hline
    Original:\textless{}Original Text\textgreater{} \\ Rewrite:\textless{}Paraphrased Text\textgreater{} \\  \hline
    \lbrack Original\rbrack:\textless{}Original Text\textgreater{} \\ \lbrack Rewrite\rbrack:\textless{}Paraphrased Text\textgreater{} \\ \hline
    Here is the original source: \textless{}Original Text\textgreater{} \\  Here is the paraphrase: \textless{}Paraphrased Text\textgreater{} \\ \hline
    \textless{}Original Text\textgreater{}, in other words \textless{}Paraphrased Text\textgreater{}\\
    \end{tabular}
    \caption{The templates that were used to give examples of how the paraphrasing should be done to the pre-trained language model.}
    \label{table:onlyexamples}
\end{table}

In instruction for prompts, we provided examples and simple instructions to the language models. The instructions were used to ask the model to generate paraphrases before presenting them with examples. Table \ref{table:examplewithinstruction} shows the instructions we used to explain the task to the model at the beginning of our prompts.

\begin{table}[!ht]
    \centering 
    \begin{tabular}{>{\centering\arraybackslash}p{3in}}
    \textbf{Instructions}                               \\ \hline
    Summarize the following text in your own words \\ \hline
    Rewrite the following text that expresses the same idea in a different way\\  \hline
    Generate a paraphrase of the following text that expresses the same ideas in a different way \\ \hline
    Generate a paraphrase of the following text using different words and sentence structures while still conveying the same meaning\\ \hline
    Generate a summary or paraphrase of the following text that captures the essence of the ideas in a concise manner\\ 
    \end{tabular}
    \caption{The instructions that were used before giving examples to the language model to describe the paraphrasing task.}
    \label{table:examplewithinstruction}
\end{table}

\section{Contrastive Prompt-based Fine-tuning Details}
\label{appendix:contrastive prompt-based fine-tuning details}

Contrastive prompt-based fine-tuning contains two main steps: (1) Masked Language Modeling and (2) Contrastive Learning.

\begin{algorithm}[!ht]
  \caption{Learning from MLM and SupCon with Paraphrasing}
  \label{alg:mlm-supcon}
  \begin{algorithmic}[1]
    \State \textbf{Input:}
    \State Training set: $\mathcal{D}_{train}$
    \State MLM model: $\mathcal{M}$
    \State Function to concatenate two strings: $Concat$
    \State Cross Entropy loss: $CE$
    \State Supervised Contrastive loss: $SupCon$
    \State Paraphrase function: $Paraphrase$ 
    \State Function that samples from a dataset and puts it in the specific template: $Sample$
    \State // The third parameter of this function specifies
    \State // whether to pus \lbrack MASK\rbrack or the verbalizer of
    \State // the label
    \State Template For Prompts: $Template$
    \State $MaxStep=1000$
    \State \textbf{Preparing Samples:}
    \For{i < MaxStep}
        \State $sent,y$=Sample($\mathcal{D}_{train}$, $Template$, false)
        \State $demo_1$=Sample($\mathcal{D}_{train}$, $Template$, true)
        \State $demo_2$=Sample($\mathcal{D}_{train}$, $Template$, true)
        \State $demo_3$=Sample($\mathcal{D}_{train}$, $Template$, true)
        \State $demo_4$=Sample($\mathcal{D}_{train}$, $Template$, true)
        \State $demo_{in_1}$=Concat($demo_1$, $demo_2$,)
        \State $demo_{in_2}$=Concat($demo_3$, $demo_4$,)
        \State $x_{in_1}$=Concat($\mathcal{T}(sent), \mathcal{T}(demo_{in_1})$)
        \State $x_{in_2}$=Concat($\mathcal{T}$(Par($sent$))$, \mathcal{T}(demo_{in_2})$)
        \State $\vartriangleright$ \textbf{MLM Learning:}
        \State $output_1$ = $\mathcal{M}(x_{in_1})$
        \State $\mathcal{L}_{MLM}$ = CE($output_1, y$)
        \State $\mathcal{L}_{MLM}$.backward()
        \State optimizer.step()
        \State $\vartriangleright$ \textbf{Contrastive Learning:}
        \State $output_2$ = $\mathcal{M}(x_{in_2})$
        \State $\mathcal{L}_{SupCon}$ = SupCon($output_1, output_2$)
        \State $\mathcal{L}_{SupCon}$.backward()
        \State optimizer.step()
    \EndFor
  \end{algorithmic}
\end{algorithm}

\paragraph{Masked Language Modeling (MLM) Loss.} A classification task is approached as a Masked Language Modeling(MLM) problem in prompt-based methods. The input consists of a sentence (sent) and a template with a mask (temp) (i.e., $x_{prompt} = sent, temp(\lbrack$MASK$\rbrack)$), and the goal is to determine the best token to fill in the $\lbrack$MASK$\rbrack$. This results in a MLM loss, represented as $\mathcal{L}_{MLM} = MLM(x_{prompt}, y)$, where $y$ is the word label associated with $x_{prompt}$. LM-BFF \cite{gao2021making} uses demonstrations of label words to improve the results. The input for this approach includes the sentence ($sent_0$) and the masked template ($temp_0$) with a mask ([MASK]. The input also contains an additional sentence ($sent_i$) with the same template ($temp_0$) with its own verbalizer ($word_i$) for those sentences. The label words are sampled from the training set. The classification loss is then calculated using this input.

The language model first encodes the input sentence $x_{in}$ into a sequence of tokens, which are then mapped to a sequence of hidden states ${ h_1, h_2, ..., h_{L} }$. $ L$ denotes the length of the sequence, and the dimension of the hidden states is denoted by $d$. For example, in prompt-based fine-tuning, if the input sentence ($x_{in}$) is ``France missed the world cup in penalties,'' the corresponding prompt $x_{prompt}$ would be $\lbrack$CLS$\rbrack$ $x_{in}$, $\lbrack$MASK$\rbrack$.$\lbrack$SEP$\rbrack$. The model then determines whether it is more likely to place the appropriate verbalizer at the [MASK] position. It has been found that fine-tuning with this fill-in-the-blank framework is superior to standard fine-tuning. The prediction of the model $\mathcal{M}$ for a class $y \in \mathcal{Y}$ can be expressed by mapping the label space Y to the label words, where $\mathcal{V}(y)$ represents the label word for class $y$. This can be written as:

\begin{equation}
        \begin{split}
        p(y|x_{in})=p(\lbrack MASK\rbrack = \mathcal{V}(y)|x_{in}) \\
        =\frac{exp(w_{\mathcal{V}(y)}.h_{\lbrack MASK\rbrack})}{\sum_{y'\in \mathcal{Y}}^{} exp(w_{\mathcal{V}(y')}.h_{\lbrack MASK\rbrack}) } 
    \end{split}
\end{equation}
where the weight vector of the MLM head is denoted by $w$.

In LM-BFF, the authors add demonstrations to the input $x_{prompt}$ to improve the model's understanding of verbalizers. As a result, the input to LM-BFF is in the following form:
\begin{equation}
    \mathcal{T}(x_{in}) \oplus \mathcal{T}(x_{in}^1,y^1) \oplus ... \oplus \mathcal{T}(x_{in}^k,y^k) 
\end{equation}

where $\mathcal{T}(x_{in}^i,y^i)$ illustrates the $i$-th demonstration in the template $mathcal{T}$ with where the actual verbalizer of the samples replaces the [MASK]. Also, $k$ is the number of demonstrations we want to use in our prompts. This paper uses random sampling to select demonstrations from the training set. The MLM loss is calculated as follows:
\begin{equation}
    \mathcal{L}_{MLM} = \sum_{(x_{in},y)\in \mathcal{D}_{train}} -log\lbrack p(y|x_{in})\rbrack 
\end{equation}

\paragraph{Supervised Contrastive Loss.}
Supervised Contrastive Learning is a specific form of contrastive learning \cite{chen2020simple, tian2020contrastive, liu2021bootstrapping} that clusters two augmented batches at the class level in the feature space and calculates the contrastive loss using Equation \ref{eq:supconloss}:
\begin{equation}
    \label{eq:supconloss}
    \mathcal{L}_{SupCon}=(x'_1, x'_2, y)
\end{equation}
where $x'_1$ and $x'_2$ are the augmented version of the input batch $x$ and $y$ is the actual label of the batch.

To use SupCon on multiple views of an input text, we first need to obtain two views of the text:
\begin{equation}
   x_{in_1}=\mathcal{T}(sent) \oplus \mathcal{T}(demo_1) \oplus \mathcal{T}(demo_2)
\end{equation}
\begin{equation}
   x_{in_2}=\mathcal{T}(Par(sent)) \oplus \mathcal{T}(demo_3) \oplus \mathcal{T}(demo_4) 
\end{equation}
where $x_{in_1}$ is the same as $x_{prompt+demo}$ in LM-BFF and $\mathcal{T}$ is a function that formats the sentence according to a specific template. Instead of using a new template in which the newly generated sample does not provide a new perspective, we use the few-shot paraphrasing ($Par$) function. Also, $verb$ stands for the verbalizer used for the actual label of the sample. Now using Equation \ref{eq:supconloss} on two views, we can calculate the total loss:
\begin{equation}
    \mathcal{L}_{Total}=\mathcal{L}_{SupCon} + \mathcal{L}_{MLM}
\end{equation}

Algorithm \ref{alg:mlm-supcon} shows an overview of our method which uses contrastive few-shot fine-tuning with few-shot paraphrasing. It is important to mention that learning from $\mathcal{L}_{SupCon}$ requires one additional forward and backward pass, which increases the computational cost by a factor of 1.5. However, the cost is still the same as \newcite{jian-etal-2022-contrastive}'s model due to the $O(1)$ time complexity of the $Paraphrase$ function. Figure \ref{fig:fine-tuning} shows the fine-tuning procedure for one prompt sample and its new view created using few-shot paraphrasing.

\end{document}